\documentclass{article} 
\usepackage{iclr2027_conference,times}

\usepackage{amsmath,amsfonts,bm}

\def\eqref#1{equation~\ref{#1}}

\def\1{\bm{1}}

\DeclareMathAlphabet{\mathsfit}{\encodingdefault}{\sfdefault}{m}{sl}
\SetMathAlphabet{\mathsfit}{bold}{\encodingdefault}{\sfdefault}{bx}{n}

\usepackage{hyperref}
\usepackage{url}

\usepackage{amsmath}
\usepackage{amssymb}
\usepackage{marvosym} 
\usepackage{mathtools}
\usepackage{amsthm}
\usepackage{listings}

\usepackage{amsmath}
\usepackage[utf8]{inputenc} 
\usepackage[T1]{fontenc}    
\usepackage{hyperref}       
\usepackage{url}            
\usepackage{booktabs}       
\usepackage{amsfonts}       
\usepackage{nicefrac}       
\usepackage{microtype}      
\usepackage{xcolor}         
\usepackage{xspace}
\usepackage{graphicx}
\usepackage{makecell}
\usepackage{colortbl}
\usepackage{tcolorbox}      
\usepackage{enumitem}       

\usepackage{xcolor}
\usepackage{tcolorbox}
\usepackage[normalem]{ulem}

\usepackage{subcaption}
\usepackage{wrapfig}
\usepackage{sidecap, caption}

\usepackage[normalem]{ulem}

\usepackage[capitalize,noabbrev]{cleveref}

\usepackage{algorithm}
\usepackage{algpseudocode}

\algrenewcommand{\algorithmiccomment}[1]{%
  \hfill$\triangleright$~#1%
}

\usepackage{mathtools}

\usepackage{bbm}

\newcommand{\method}{\text{VOPD}\xspace}

\definecolor{fancyblue}{rgb}{0.0, 0.45, 0.81}

\title{Verifiable On-Policy Distillation}

\author{Juntong Shi\thanks{Corresponding author. \Letter\ \texttt{juntong@stanford.edu}.}, Stefano Ermon, Jure Leskovec, Minkai Xu\\
Stanford University\\
}

\iclrfinalcopy 
\begin{document}

\maketitle

\begin{abstract}
Reinforcement learning with verifiable rewards (RLVR) has become a powerful approach for improving language-model reasoning, but its supervision is inherently sparse: a sequence-level verifier reward is propagated only through sampled trajectories and provides no direct signal for alternative actions. On-policy distillation (OPD) offers a complementary form of supervision by providing dense teacher feedback over the full next-token distribution, but this signal is not grounded in verified task outcomes. Simply co-optimizing the two objectives does not couple these signals: verifier feedback remains sparse and does not inform the dense teacher supervision. We introduce Verifiable On-Policy Distillation (\method), which transfers sparse verifier feedback into dense, verifier-grounded supervision. Our key observation is that the implicit-reward view of OPD places teacher supervision and verifier feedback in a common reward space while retaining a reward signal available for every next-token action. Under this view, \method projects verifier preference onto the teacher's implicit-reward direction and Rao--Blackwellizes the resulting signal over the full vocabulary, propagating verifier-grounded supervision to actions that were never sampled. The resulting objective can also be interpreted as adaptively extrapolating the distillation target according to teacher--verifier agreement. Across two teacher--student configurations and eight mathematical reasoning benchmarks, \method consistently improves both final performance and learning efficiency over vanilla OPD and other on-policy distillation baselines. Our code is available at \url{https://github.com/juntongshi48/VOPD}.
\end{abstract}

\section{Introduction}
Recent advances in reinforcement learning with verifiable rewards (RLVR) have substantially improved language-model reasoning~\citep{deepseekmath,deepseekr1}, especially in domains such as mathematics and code where outcomes can be checked automatically. However, its supervision is inherently sparse: a sequence-level verifier reward is propagated only through sampled trajectories and provides no direct signal for alternative actions~\citep{rewarding-progress,rlsd}. As a result, verified feedback is well grounded in task outcomes but makes inefficient use of each model-generated trajectory~\citep{rewarding-progress}.

On-policy distillation (OPD) provides a complementary form of supervision~\citep{opd-iclr,lu2025onpolicydistillation}. By querying a stronger teacher at student-visited prefixes, OPD provides dense distributional feedback over the full next-token vocabulary, including unsampled actions~\citep{rethinking,sdpo}, enabling substantially more learning from each generated trajectory. Yet this signal reflects the teacher's preferences rather than verified task outcomes; when teacher preference and verifier feedback disagree, OPD can reinforce actions that do not improve task success~\citep{opdvr}.

These complementary strengths suggest a promising direction: combining RLVR and OPD to obtain supervision that is both verifier-grounded and dense~\citep{rlsd,opdvr}. However, simply optimizing the two objectives together does not achieve this~\citep{opdvr,safopd}. The teacher signal remains independent of verified outcomes and can therefore conflict with the verifier, while the verifier signal remains restricted to sampled trajectories and does not reshape the dense supervision over alternative actions. Consequently, directly adding verifier-based RL to OPD yields only marginal improvements~\citep{opdvr,safopd}. Recent methods have explored tighter combinations of teacher and verifier feedback~\citep{rlsd,rlcsd,opdvr,uni-opd}, but either restrict verifier-guided supervision to sampled actions or use predefined reweighting rules to modulate the dense teacher signal. What is missing is a principled representation through which sparse verifier feedback can be converted into dense action-level supervision.

To address this, we introduce \method, which transfers sparse verifier feedback into dense, verifier-grounded supervision through the teacher's implicit reward. Our key observation is that the implicit-reward view of OPD provides exactly the representation needed for this transfer. Reverse-KL distillation toward a teacher can equivalently be viewed as maximizing the teacher's implicit reward
\[
d(s,a)
=
\log\frac{p_T(a\mid s)}{p_{\mathrm{ref}}(a\mid s)}
\]
under KL regularization to a reference policy~\citep{dpo,implicitprm,gopd}. This view places teacher supervision and verifier feedback in a common reward space while retaining a crucial advantage of OPD: the teacher's implicit reward can be evaluated for every next-token action at each prefix. In \method, we exploit this implicit-reward view to transfer verifier preference into the dense teacher-reward signal. For each prompt, we accumulate the teacher's implicit rewards along student-generated responses and compare them with their verifier outcomes. We then project the verifier advantage onto the teacher-reward direction, extracting the component of verifier preference that can be represented through the teacher's implicit reward. Because the projected verifier signal remains a scalar reweighting of the token-wise implicit reward $d(s,a)$, it can be evaluated for every action in the vocabulary. We can therefore Rao--Blackwellize the corresponding sampled-action estimator over the full vocabulary, yielding dense, verifier-grounded supervision without additional verifier evaluations.
The resulting objective also admits an equivalent adaptive-extrapolation interpretation. Rather than distilling toward a fixed teacher distribution, \method shifts the target along the teacher--reference log-probability direction by an amount determined by teacher--verifier agreement. Positive agreement strengthens movement along the teacher-reward direction, while weak or negative agreement suppresses or reverses it.

Empirically, we evaluate \method across two teacher--student settings and eight mathematical reasoning benchmarks, covering both same-size and strong-to-weak distillation. Compared with other on-policy distillation methods, \method consistently improves both final performance and learning efficiency. These results support our central premise: sparse verifier feedback can be substantially more effective when transferred through the teacher's implicit-reward representation than when optimized as a separate sparse objective.

\begin{figure*}[!t]
    \centering
    \includegraphics[width=0.99\linewidth]{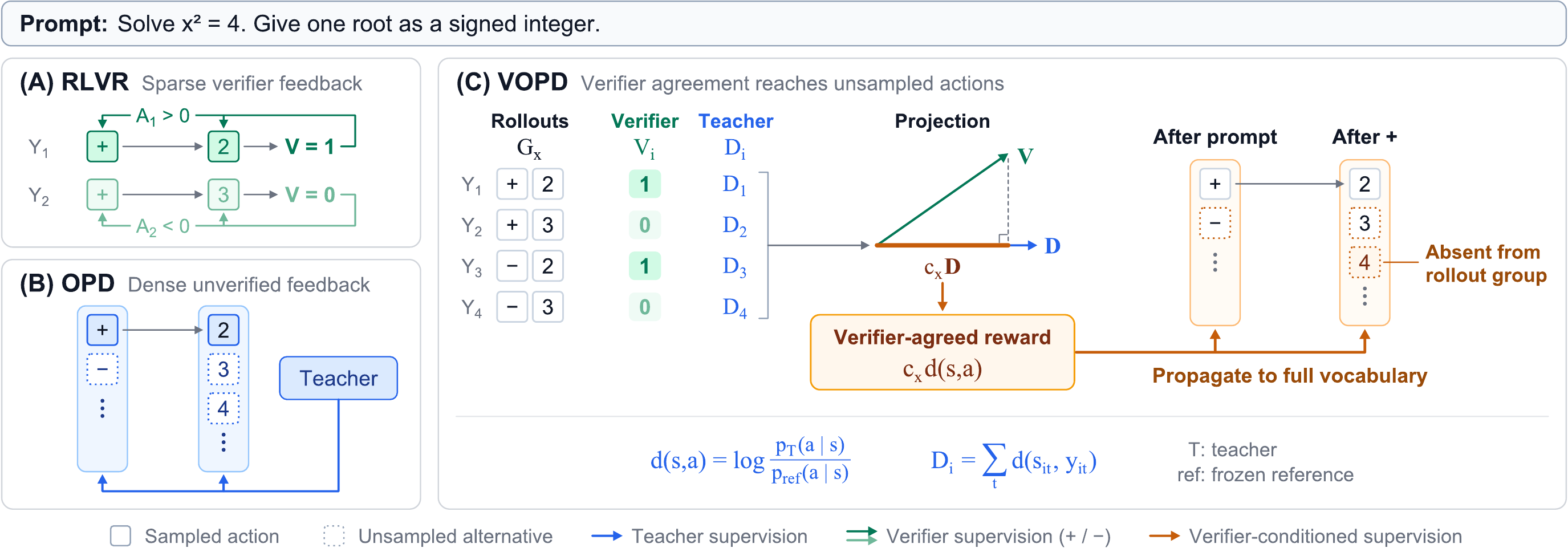}
    \caption{
        Comparison of RLVR, OPD, and \method. 
        \textbf{(A)} RLVR assigns a sparse but verified to the sampled tokens only. 
        \textbf{(B)} OPD provides dense teacher feedback over the full next-token distribution, but this feedback is not grounded in verified outcomes. 
        \textbf{(C)} \method projects verifier preference onto the teacher's implicit-reward direction, then Rao--Blackwellizes the resulting signal over hypothetical next-token actions to obtain dense, verifier-grounded supervision across full vocabulary.
    }
    \label{fig:vopd-main}
    \vspace{-5pt}
\end{figure*}

\section{Related Work}

\textbf{Reinforcement Learning with Verifiable Rewards (RLVR).}
RLVR has become a widely used paradigm for improving language-model reasoning by optimizing directly against automatically checkable outcomes~\citep{deepseekmath,deepseekr1}. Early RL-based post-training commonly builds on Proximal Policy Optimization (PPO)~\citep{ppo}, while more recent reasoning systems often use variants such as GRPO that avoid training a separate value model. These methods use verifier-derived sequence-level advantages to reinforce successful trajectories. While effective, this supervision is inherently sparse: verifier feedback is available only for sampled responses and is typically shared across all tokens within a trajectory.

\textbf{On-Policy Distillation (OPD).}
On-policy distillation transfers the behavior of a stronger teacher into a weaker, often smaller, student~\citep{opd-iclr,lu2025onpolicydistillation}. OPD minimizes the reverse KL from the student distribution to the teacher distribution on trajectories generated by the student itself, making the optimization on-policy in a manner analogous to RLVR. When teacher logits are available, this objective provides dense supervision over the full next-token vocabulary at each visited prefix rather than only the sampled action.
A growing line of work~\cite{gopd,privileged-information-distillation} studies how to improve or reinterpret this dense distillation signal. In particular, G-OPD~\citep{gopd} views reverse-KL OPD through the teacher's implicit reward relative to a reference policy and extrapolates along this teacher--reference direction with a fixed strength. Our method builds on the same implicit-reward view, but uses verifier feedback to adapt how strongly this direction is followed for each prompt group rather than relying on a fixed extrapolation factor.

\textbf{Combining Distillation with Verifier Feedback.}
Several recent methods combine teacher supervision with verifier-grounded learning. RLSD and RLCSD~\citep{rlsd,rlcsd} use verifier outcomes to determine the direction of the policy update, while teacher--student likelihood differences provide finer-grained token-level credit magnitudes. Although this improves credit assignment within a sampled trajectory, the verifier-guided signal remains tied to the sampled actions rather than extending to the full vocabulary.
A complementary line of work preserves the dense teacher direction while using verifier outcomes to modulate its strength. OPDVR~\citep{opdvr} gates teacher-derived updates according to trajectory correctness, while Uni-OPD~\citep{uni-opd} adjusts teacher rewards based on the relative teacher scores of correct and incorrect rollouts. These constructions provide intuitive ways to incorporate outcome feedback into dense distillation, but rely on predefined rules for translating sequence-level verifier outcomes into modifications of the teacher signal. In contrast, \method derives this interaction directly by projecting verifier preference onto the teacher's implicit-reward direction and extends the resulting signal to full-vocabulary supervision without changing its expected local update.

Another popular way to exploit verifier feedback is through self-distillation~\citep{opsd,sdpo,sdft}, where the student bootstraps itself by using a hindsight-informed copy as the teacher. These methods incorporate verifier outcomes or feedback as additional context to the teacher-side prefix, then distill the resulting teacher logits back into the original student. Although these methods also leverage verifier feedback, they do so along a largely orthogonal axis to \method: rather than changing how verifier feedback reshapes a fixed teacher signal, they change the teacher distribution through hindsight context. In principle, \method can be applied in the same self-distillation setting to further improve how this hindsight-informed teacher signal is transferred.

\section{Preliminaries}
\label{sec:method:preliminaries}

Let $p_\theta$ denote the student policy, $p_T$ the teacher, and $p_{\mathrm{ref}}$ a fixed reference policy, all defined over a shared vocabulary $\mathcal V$. For each prompt $x$, we sample a group of $n$ responses $\mathcal G_x=\{Y_i\}_{i=1}^n$ from the current student, $p_\theta$. We write $s_{it}=(x,y_{i,<t})$ for the prefix preceding token $y_{it}$ and $V_i=V(x,Y_i)$ for the verifier reward assigned to response $Y_i$.

\textbf{Verifier-Based Policy Optimization.}
A direct way to align a policy with verifiable task outcomes is policy-gradient reinforcement learning. A simple and widely used instance is Group Relative Policy Optimization (GRPO), which samples multiple responses to the same prompt and constructs an advantage by comparing their verifier rewards~\citep{deepseekmath}. Specifically, with group mean
$
\bar V
=
\frac{1}{n}\sum_{i=1}^n V_i
$
, and standard deviation
$
\sigma_V
=
\sqrt{\frac{1}{n}\sum_{i=1}^n(V_i-\bar V)^2}
$
, for groups with $\sigma_V>0$, GRPO defines the group-relative verifier advantage
\begin{equation}
A_i^V
=
\frac{V_i-\bar V}{\sigma_V}.
\label{eq:verifier-advantage}
\end{equation}
Suppressing clipping and importance-weighting details, the corresponding policy-gradient objective takes the form
\begin{equation}
\mathcal L_{\mathrm{GRPO}}
=
-
\mathbb E_{x,i,t}
\left[
A_i^V
\log p_\theta(y_{it}\mid s_{it})
\right].
\label{eq:grpo-simplified}
\end{equation}
The resulting supervision is grounded in verified task outcomes, but remains sparse: it is available only for sampled responses, and the same sequence-level advantage is applied uniformly to all sampled steps in each trajectory.

\textbf{On-Policy Distillation and Implicit Reward.}
On-policy distillation provides a complementary way to transfer the behavior of a stronger teacher into a student. At each prefix $s$ visited by the student, OPD minimizes the reverse-KL divergence
\begin{equation}
\ell_{\mathrm{OPD}}(\theta;s)
=
D_{\mathrm{KL}}\!\left(
p_\theta(\cdot\mid s)
\,\|\,p_T(\cdot\mid s)
\right),
\label{eq:opd}
\end{equation}
providing distributional supervision over the entire next-token vocabulary~\citep{opd-iclr,lu2025onpolicydistillation}.

To place OPD in a reward-based form, define the teacher's implicit reward relative to the reference as
$
d(s,a)
=
\log
\frac{p_T(a\mid s)}
{p_{\mathrm{ref}}(a\mid s)}.
$
The reverse-KL objective satisfies
\begin{equation}
\begin{aligned}
\ell_{\mathrm{OPD}}(\theta;s)
&=
D_{\mathrm{KL}}\!\left(
p_\theta(\cdot\mid s)
\,\|\,p_{\mathrm{ref}}(\cdot\mid s)
\right)
-
\mathbb E_{a\sim p_\theta(\cdot\mid s)}
\left[d(s,a)\right].
\end{aligned}
\label{eq:opd-reward-identity}
\end{equation}
Thus, reverse-KL OPD can equivalently be viewed as maximizing the teacher's implicit reward $d(s,a)$ under KL regularization to the reference~\citep{free-process-reward,gopd}. Unlike verifier feedback, which is available only on sampled trajectories, $d(s,a)$ can be evaluated for every $a\in\mathcal V$ at each visited prefix, providing dense supervision over alternative actions. This implicit-reward view therefore puts OPD and verifier-based RL in a common reward-based language.

\section{Verifiable On-Policy Distillation}
\label{sec:method}

We introduce \method, which converts sparse verifier feedback into verifier-grounded dense supervision using the teacher's implicit reward. Starting from the additive combination of OPD and verifier-based policy optimization, \method projects the verifier advantage onto the teacher implicit-reward direction, maps the projected signal back to the scale of the implicit reward, and Rao--Blackwellizes the resulting sampled-action signal over the full vocabulary.

\subsection{Beyond Co-Optimization}
\label{sec:method:setup}

A direct combination of OPD and RLVR is to co-optimize both objectives
\begin{equation}
\mathcal L_{\mathrm{add}}
=
\mathbb E_{x,s}
\left[
\ell_{\mathrm{OPD}}(\theta;s)
\right]
+
\lambda\mathcal L_{\mathrm{GRPO}},
\label{eq:additive}
\end{equation}
where $\lambda\geq0$ controls the strength of verifier-based supervision relative to ordinary OPD. However, simply adding the GRPO objective has limited effect because its supervision is substantially sparser and coarser than the dense distributional signal provided by OPD. This discrepancy is also reflected empirically: verifier-only RL requires far more optimization steps to reach comparable accuracy, and directly adding GRPO yields only marginal gains over vanilla OPD.

The implicit-reward formulation of OPD in~\Cref{eq:opd-reward-identity} suggests a natural way to integrate the two signals more tightly. Instead of keeping verifier feedback as a separate sparse objective, we express its teacher-representable component through the teacher's implicit reward. This allows verifier information to inherit OPD's dense action-level structure.

\subsection{Projecting Verifier Feedback onto the Teacher's Implicit Reward}
\label{sec:method:projection}

The verifier reward $V_i$ is defined for an entire response, whereas the teacher's implicit reward $d(s,a)$ is defined for individual token actions. To place them at the same level, we aggregate the latter along each sampled response:
\begin{equation}
D_i
=
\sum_{t=1}^{|Y_i|}
d(s_{it},y_{it})
=
\log
\frac{p_T(Y_i\mid x)}
{p_{\mathrm{ref}}(Y_i\mid x)}.
\label{eq:teacher-sequence-reward}
\end{equation}
We refer to $D_i$ as the cumulative teacher's implicit reward. Crucially, this sum is itself an exact response-level implicit reward: if $Y_i$ is treated as a single action, $D_i$ measures the teacher's preference for the entire response relative to the reference. 

Each sampled response is therefore associated with two response-level signals: the verifier reward $V_i$ and the cumulative teacher's implicit reward $D_i$. Because the two rewards can differ in offset and scale, we compare them through standardized within-group advantages. GRPO already defines
$A_i^V$ in \Cref{eq:verifier-advantage}; analogously, we standardize the cumulative teacher rewards within each prompt group:
\begin{equation}
A_i^D
=
\frac{D_i-\bar D}{\sigma_D},
\label{eq:teacher-advantage}
\end{equation}
where $\bar D$ and $\sigma_D$ denote the group mean and standard deviation of $\{D_i\}_{i=1}^n$.

Centering removes prompt-specific offsets, while standardization removes arbitrary differences in reward scale. Both transformations preserve the within-group ordering of responses, so $\mathbf A^V$ and $\mathbf A^D$ describe the relative preferences induced by the verifier and teacher over alternative attempts at the same prompt on a common scale.

We seek to transfer verifier preference through the teacher's implicit reward. To this end, we restrict the transferred signal to scalar multiples of the teacher-reward direction, $b\,\mathbf A^D$ for $b\in\mathbb R$. This restriction is what makes dense transfer possible: scaling the sequence-level teacher reward by $b$ scales each additive token reward $d(s_{it},y_{it})$ by the same factor. The resulting rescaled sampled-action signal can therefore easily be extended later to the full vocabulary.

Within this family, we choose the signal that best matches the observed verifier advantages:
\begin{equation}
c_x
=
\arg\min_b
\frac{1}{n}
\sum_{i=1}^n
\left(
A_i^V-bA_i^D
\right)^2.
\label{eq:projection-problem}
\end{equation}
The solution is
\begin{equation}
\begin{aligned}
c_x
=
\frac{1}{n}
\sum_{i=1}^n
A_i^V A_i^D
=
\frac{
\frac{1}{n}
\sum_i
(V_i-\bar V)(D_i-\bar D)
}{
\sigma_V\sigma_D
}
\in[-1,1],
\end{aligned}
\label{eq:alignment}
\end{equation}
and the projected verifier advantage is
\begin{equation}
\widehat A_i^V
=
c_xA_i^D.
\label{eq:projected-verifier-advantage}
\end{equation}
$\widehat A_i^V$ is the component of verifier preference that can be represented through the teacher-reward direction. Since both $A_i^V$ and $A_i^D$ are standardized, the projection factor $c_x$ is the empirical correlation between verifier reward and cumulative teacher's implicit reward within the prompt group. Positive $c_x$ indicates that responses preferred more strongly by the teacher also tend to achieve better verified outcomes, so the teacher-reward direction should be reinforced; negative $c_x$ indicates that verified outcomes favor movement against that direction, so it should instead be opposed.

\subsection{From Projected Verifier Feedback to Full-Vocabulary Supervision}
\label{sec:method:dense}
The projected verifier advantage
$\widehat A_i^V=c_xA_i^D$
is expressed in standardized advantage coordinates. Substituting the definition of $A_i^D$ gives
\begin{equation}
\widehat A_i^V
=
\frac{c_x}{\sigma_D}(D_i-\bar D).
\label{eq:projected-advantage}
\end{equation}
Using this quantity directly would make the effective strength of the transferred signal scale inversely with $\sigma_D$. We therefore rescale it by $\sigma_D$, placing it on the same scale as the cumulative teacher reward:
\begin{equation}
\sigma_D\widehat A_i^V
=
c_x(D_i-\bar D).
\label{eq:transferred-sequence-reward}
\end{equation}
The transferred signal involves the cumulative teacher reward $D_i=\sum_t d(s_{it},y_{it})$. Directly using this sequence-level reward in a token-level policy gradient couples each action's update to future rewards, which cannot be evaluated for every candidate action without considering its possible continuations. Following the immediate-token approximation used in OPD~\citep{revisiting-opd}, we instead replace the future reward return with the current-token implicit reward $d(s,y)$, treating each visited prefix $s$ as fixed. This yields the local sampled-action estimator
\begin{equation}
\widehat g(s)
=
c_x\bigl(d(s,a)-\bar D\bigr)
\nabla_\theta\log p_\theta(a\mid s).
\label{eq:local-estimator-with-baseline}
\end{equation}
Crucially, this signal need not remain restricted to the sampled action.
At every visited prefix, the teacher and reference provide
$
d(s,a)
=
\log\frac{p_T(a\mid s)}{p_{\mathrm{ref}}(a\mid s)}
$
for every $a\in\mathcal V$. We can therefore integrate over the next-token distribution.
Specifically, we treat the rollout group used to compute $c_x$ and the visited prefix $s$ as already observed, and consider only randomness from sampling new hypethetical next token. This lead to the definition of the Rao--Blackwellized estimator as the conditional expectation of $\widehat g$ given $\mathcal G_x$ and $s$ in the estimator notation:
\begin{equation}
\begin{aligned}
\widetilde g(s)
&:=
\mathbb E_{a\sim p_\theta(\cdot\mid s)}
\left[
\widehat g(s,a)\mid\mathcal G_x,s
\right]
\end{aligned}
\label{eq:rao-blackwell-gradien-definition}
\end{equation}
By construction, $\widetilde g(s)$ is the conditional expectation of the
local sampled-action estimator. Hence, by the law of total expectation,
$
\mathbb E_{\mathcal G_x,s}[\widetilde g(s)]
=
\mathbb E_{\mathcal G_x,s,a}[\widehat g(s,a)]
$, meaning that Rao-Blackwellization preserves the expected local gradient after adopting the next-token approximation.  Further simplifying $\widetilde g(s)$ gives
\begin{equation}
\begin{aligned}
\widetilde g(s)
&=
\mathbb E_{a\sim p_\theta(\cdot\mid s)}
\left[
c_x\bigl(d(s,a)-\bar D\bigr)
\nabla_\theta\log p_\theta(a\mid s)
\mid\mathcal G_x,s
\right]\\
&=
c_x
\mathbb E_{a\sim p_\theta(\cdot\mid s)}
\left[d(s,a)
\nabla_\theta\log p_\theta(a\mid s)
\mid\mathcal G_x,s
\right]\\
&=
c_x
\sum_{a\in\mathcal V}
d(s,a)\nabla_\theta p_\theta(a\mid s)\\
&=
\nabla_\theta
\left[
c_x
\sum_{a\in\mathcal V}
p_\theta(a\mid s)d(s,a)
\right].
\end{aligned}
\label{eq:rao-blackwell-gradient-derivation}
\end{equation}
Since $G_x$ and $s$ are already observed and fix, $c_x$ is fixed with respect to new hypethetical next-token sample $a$, and the baseline term $\bar D$ vanishes by the score-function identity $\mathbb E_{a\sim p_\theta(\cdot\mid s)}[\nabla_\theta\log p_\theta(a\mid s)]=0$. These observations justify the second second equality.

Thus, Rao-Blackwellization preserves the
expected local gradient while integrating out next-token sampling variance, and in doing so turns the verifier-derived signal from a sampled-action update into dense, verifier-grounded supervision over the full vocabulary without additional verifier evaluations.

Moreover, Equation~\eqref{eq:rao-blackwell-gradient-derivation} shows that this
Rao--Blackwellized estimator is exactly the gradient of the
full-vocabulary objective
\begin{equation}
j_{\mathrm{VOPD}}(\theta;s)
=
c_x
\sum_{a\in\mathcal V}
p_\theta(a\mid s)d(s,a).
\label{eq:dense-vopd-reward}
\end{equation}

\begin{algorithm}[t]
\caption{\method: Verifiable On-Policy Distillation}
\label{alg:vopd}
\begin{algorithmic}[1]
\Require Student $p_\theta$, teacher $p_T$, reference $p_{\mathrm{ref}}$,
training question set $\mathcal D=\{x_j\}_{j=1}^N$, verifier $V$,
group size $n$, verifier weight $\lambda$

\For{each training iteration}
    \State Sample a batch of prompts $x$ from $\mathcal D$

    \For{each prompt $x$}
        \State \textcolor{fancyblue}{\textit{// Step 1: On-policy rollout}}
        \State Sample $n$ responses $\{Y_i\}_{i=1}^n \sim p_\theta(\cdot\mid x)$
        \State Obtain verifier rewards $V_i = V(x,Y_i)$

        \State \textcolor{fancyblue}{\textit{// Step 2: Teacher--verifier projection}}
        \For{$i=1,\dots,n$}
            \State Evaluate $p_T$ and $p_{\mathrm{ref}}$ on prefixes of $Y_i$
            \State $D_i \gets
            \sum_t
            \log
            \frac{p_T(y_{it}\mid s_{it})}
                 {p_{\mathrm{ref}}(y_{it}\mid s_{it})}$
        \EndFor
        \State Compute
        $A_i^V=(V_i-\bar V)/\sigma_V$
        and
        $A_i^D=(D_i-\bar D)/\sigma_D$
        \State $c_x \gets \frac{1}{n}\sum_{i=1}^n A_i^V A_i^D$
        \Comment{Projection coefficient}
    \EndFor

    \State \textcolor{fancyblue}{\textit{// Step 3: Verifiable distillation update}}
    \State Compute the gradient ascent direction
    \[
    \begin{aligned}
    g_{\mathrm{\method}}
    =
    \mathbb E_{x,s}
    \left[
    (1+\lambda c_x)
    \sum_{a\in\mathcal V}
    p_\theta(a\mid s)d(s,a)
    \nabla_\theta\log p_\theta(a\mid s)
    -
    \nabla_\theta D_{\mathrm{KL}}
    \left(p_\theta(\cdot\mid s)\|p_{\mathrm{ref}}(\cdot\mid s)\right)
    \right]
    \end{aligned}
    \]
    \State Update $\theta \gets \theta + \eta g_{\mathrm{\method}}$
\EndFor
\end{algorithmic}
\end{algorithm}

\subsection{The \method Objective}
\label{sec:method:objective}

We can now return to the co-optimization objective in Equation~\eqref{eq:additive}. Rather than retaining the sparse GRPO term, \method replaces it with the projected verifier signal derived above and Rao--Blackwellized it over the full vocabulary. Combining Equation~\eqref{eq:dense-vopd-reward} with the ordinary OPD loss gives
\begin{equation}
\begin{aligned}
\ell_{\mathrm{\method}}(\theta;s)
&=
D_{\mathrm{KL}}\!\left(
p_\theta(\cdot\mid s)
\,\|\,p_T(\cdot\mid s)
\right)\\
&\quad-
\lambda c_x
\sum_{a\in\mathcal V}
p_\theta(a\mid s)
\log
\frac{p_T(a\mid s)}
{p_{\mathrm{ref}}(a\mid s)}.
\end{aligned}
\label{eq:vopd-local-loss}
\end{equation}
The full training objective is therefore
$
\mathcal L_{\mathrm{\method}}(\theta)
=
\mathbb E_{x,s}
\left[
\ell_{\mathrm{\method}}(\theta;s)
\right]
$.

\textbf{Adaptive Extrapolation View.}
The objective also admits an equivalent target-distribution interpretation. Up to a normalization constant independent of $\theta$,~\Cref{eq:vopd-local-loss} is also equivalent to the reverse-KL distillation toward the teacher-extrapolation target
\begin{equation}
q_x(a\mid s)
\propto
p_T(a\mid s)^{1+\lambda c_x}
p_{\mathrm{ref}}(a\mid s)^{-\lambda c_x}.
\label{eq:vopd-target}
\end{equation}
Under this view, $\lambda c_x$ thus controls how far the distillation target moves along the teacher--reference log-probability direction: a positive $c_x$ extrapolates beyond the teacher, while a negative $c_x$ correspondingly moves the target back toward the reference. This connects \method to ExOPD~\citep{gopd}: ExOPD uses a fixed positive extrapolation strength regardless of verifier feedback, whereas \method adapts the extrapolation strength for each prompt group according to the observed teacher--verifier agreement.

\textbf{Zero-variance handling.}
When the verifier rewards have zero within-group variance, the correlation $c_x$ is undefined. In these cases, we fall back to a fixed extrapolation coefficient $\lambda=0.25$, while groups with defined correlation use the verifier-conditioned coefficient $\lambda c_x$. This fallback is motivated by the empirical benefit of fixed extrapolation over standard OPD, as demonstrated by ExOPD in~\Cref{sec:experiments:main-results}.

\textbf{Computational Overhead.}
\Cref{alg:vopd} summarizes the full training procedure. Compared with ordinary OPD, \method additionally requires a frozen reference model and evaluates its logits on the sampled prefixes, introducing extra model memory and one additional forward evaluation over each rollout batch. Beyond this reference-model evaluation, the remaining computation is lightweight: forming the cumulative teacher rewards and group-level coefficients $c_x$ requires only simple reductions over already-computed quantities, with no additional rollouts or verifier evaluations. Since on-policy training is typically dominated by rollout generation, the extra reference evaluation contributes only a small amount of computational overhead.

\section{Experiments}
\label{sec:experiments}
\subsection{Experiment Setup}
\label{sec:experiments:experiment-setup}
\textbf{Distillation Settings.}
We focus our experiments on mathematical reasoning and evaluate two teacher--student configurations spanning both same-size and strong-to-weak distillation. For the same-size setting, we use the RL-trained Qwen3-4B~\citep{qwen3} from ~\citet{gopd} as the teacher and distill it into Qwen3-4B non-thinking mode. For the strong-to-weak setting, we distill the same teacher into the smaller Qwen3-1.7B model. We set the reference policy $p_{\mathrm{ref}}$ to the student's initialization and keep it frozen throughout training.

\textbf{Baselines.}
We compare against three on-policy distillation baselines. \textbf{OPD} performs standard reverse-KL on-policy distillation on student-generated prefixes. \textbf{OPD+GRPO} co-optimizes the group-relative verifier objective with the OPD objective, corresponding to the naive combination in~\Cref{eq:additive}. \textbf{Ex-OPD}~\citep{gopd} extrapolates along the teacher--reference direction with a fixed strength, and can be viewed as a verifier-agnostic special case of our method with $c_x=1$.

\textbf{Training Setting.}
All methods are trained on DAPO-Math-17k~\citep{dapo} using student-generated on-policy rollouts. We a constant learning rate and one optimization update per rollout batch. All distillation methods employ the top-$k$ distillation approximation: optimizing over the student's top-$16$ tokens at each prefix. For ExOPD and our \method, we set $\lambda=0.25$ unless otherwise specified. We train each run until validation accuracy converges. We implement all methods in VERL~\citep{verl} with FSDP~\citep{fsdp} and use vLLM~\citep{vllm} for rollout generation and offline evaluation. Each run uses 8 NVIDIA A100 GPUs.

\textbf{Evaluation.}
We evaluate on eight mathematical reasoning benchmarks: AIME 2024~\citep{aime2024}, AIME 2025~\citep{aime2025}, AMC 2023~\citep{amc2023}, HMMT February 2025~\citep{hmmt25}, HMMT November 2025~\citep{hmmt25}, MATH-500~\citep{math500}, Minerva Math~\citep{minerva-math}, and OlympiadBench~\citep{olympiadbench}. We report \emph{avg@$n$}, the average correctness of independent generations for each problem. All methods are evaluated from their final training checkpoint.

For additional implementation details, including hyperparameter choices, please refer to~\Cref{app:implementation-details}.

\begin{table*}[t!]
  \centering
  \setlength{\tabcolsep}{6pt}
  \caption{\textbf{Performance under 4B same-size distillation.} The teacher is the RL-trained Qwen3-4B and the student is Qwen3-4B. Best performance is \textbf{bolded}.}  
  \resizebox{\linewidth}{!}{
    \begin{tabular}{l|ccccccccc}
    \toprule
      \textbf{Method}
      & \textbf{AIME24}
      & \textbf{AIME25}
      & \textbf{AMC23}
      & \textbf{HMMT25-Feb}
      & \textbf{HMMT25-Nov}
      & \textbf{MATH-500}
      & \textbf{Minerva}
      & \textbf{Olympiad-Bench}
      & \textbf{Average} \\
    \midrule
    Teacher
      & 72.50
      & 62.29
      & 95.00
      & 37.71
      & 43.75
      & 94.25
      & 41.64
      & 66.17
      & 64.16 \\
    Student
      & 20.83
      & 20.21
      & 6.41
      & 11.46
      & 7.50
      & 82.25
      & 30.51
      & 48.63
      & 35.97 \\
    \midrule
    OPD
      & 69.58
      & 65.83
      & \textbf{96.09}
      & 39.58
      & 46.04
      & 94.05
      & 40.44
      & 66.06
      & 64.71 \\
    OPD + GRPO
      &71.04
      &63.12
      &95.47
      &40.00
      &47.50
      &93.90
      &40.62
      &66.32
      &64.75 \\
    ExOPD
      & 70.62
      & 66.04
      & 95.62
      & 39.79
      & 48.54
      & 94.05
      & 40.44
      & \textbf{66.88}
      & 65.25 \\
    \method
      & \textbf{73.33}
      & \textbf{68.12}
      & 95.94
      & \textbf{41.04}
      & \textbf{50.42}
      & \textbf{94.55}
      & \textbf{40.99}
      & 66.65
      & \textbf{66.38} \\
    \bottomrule
    \end{tabular}
  }
  \vspace{-6pt}
  \label{tab:main_results}
\end{table*}

\begin{table*}[t!]
  \centering
  \setlength{\tabcolsep}{6pt}
  \caption{\textbf{Performance under strong-to-weak distillation.} The teacher is the RL-trained Qwen3-4B and the student is Qwen3-1.7B. Best performance is \textbf{bolded}.}
  \resizebox{\linewidth}{!}{
    \begin{tabular}{l|ccccccccc}
    \toprule
      \textbf{Method}
      & \textbf{AIME24}
      & \textbf{AIME25}
      & \textbf{AMC23}
      & \textbf{HMMT25-Feb}
      & \textbf{HMMT25-Nov}
      & \textbf{MATH-500}
      & \textbf{Minerva}
      & \textbf{Olympiad-Bench}
      & \textbf{Average} \\
    \midrule
    Teacher
      & 72.50
      & 62.29
      & 95.00
      & 37.71
      & 43.75
      & 94.25
      & 41.64
      & 66.17
      & 64.16 \\
    Student
      & 12.92
      & 13.33
      & 44.84
      & 6.46
      & 4.58
      & 72.25
      & 20.31
      & 37.61
      & 31.01 \\
    \midrule
    OPD
      & 50.62
      & 40.00
      & 86.88
      & 21.25
      & 26.25
      & 88.60
      & 34.83
      & 58.64
      & 50.88 \\
    OPD + GRPO
      & 49.58
      & 42.50
      & 85.31
      & 21.88
      & 26.67
      & 89.65
      & 33.18
      & 59.01
      & 50.97 \\

    ExOPD
      & 52.71
      & 41.04
      & \textbf{89.22}
      & \textbf{22.50}
      & 28.54
      & 88.95
      & 35.57
      & 60.20
      & 52.34 \\
    \method
      & \textbf{57.29}
      & \textbf{42.71}
      & 88.28
      & 22.29
      & \textbf{29.38}
      & \textbf{90.20}
      & \textbf{36.03}
      & \textbf{60.39}
      & \textbf{53.32} \\
    \bottomrule
    \end{tabular}
  }
  \label{tab:strong_to_weak_results}
\end{table*}

\subsection{Main Results}
\label{sec:experiments:main-results}

\textbf{\method Consistently Improves Distillation Performance.}
We first compare the converged performance of different on-policy distillation objectives. As shown in~\Cref{tab:main_results}, under 4B same-size distillation, \method achieves the strongest average performance across the eight mathematical reasoning benchmarks, improving the average accuracy from $64.71$ for vanilla OPD to $66.38$ and obtaining the best result on six of the eight benchmarks. Notably, the distilled student also exceeds the teacher's average performance ($64.16$) and the fixed-extrapolation ExOPD ($65.25$), indicating that verifier-grounded dense supervision can improve beyond direct imitation of the teacher. In contrast, co-optimizing OPD with GRPO yields only marginal improvement over vanilla OPD, suggesting that simply adding sparse verifier supervision does not provide the same benefit.

The gains persist under the more challenging strong-to-weak setting. As shown in~\Cref{tab:strong_to_weak_results}, \method achieves the highest average performance of $53.32$ and the highest accuracy on six of eight benchmarks. Although the trained 1.7B student still remains below the 4B teacher, \method consistently improves the effectiveness of transferring the teacher's reasoning capability to the smaller model. Together, these results show that the benefit of \method extends beyond same-size distillation and remains effective under substantial student--teacher capacity mismatch.

\textbf{\method Improves Learning Efficiency.}
Beyond final accuracy, \method also improves more rapidly during training. \Cref{fig:val-curve-4b} shows validation accuracy on AIME 2024 and AIME 2025 throughout 4B same-size distillation. Compared with vanilla OPD and fixed-extrapolation ExOPD, \method separates from the baselines early in training and maintains stronger validation performance through convergence. This indicates that verifier-grounded dense supervision not only improves the final distilled model, but also makes more effective use of each on-policy rollout during training.

\subsection{Ablations and Analyses}
\label{sec:experiemnts:abaltions}
\begin{figure*}[!t]
\centering
\begin{minipage}[t]{0.65\linewidth}
\vspace{0pt}
\centering
\includegraphics[width=\linewidth]{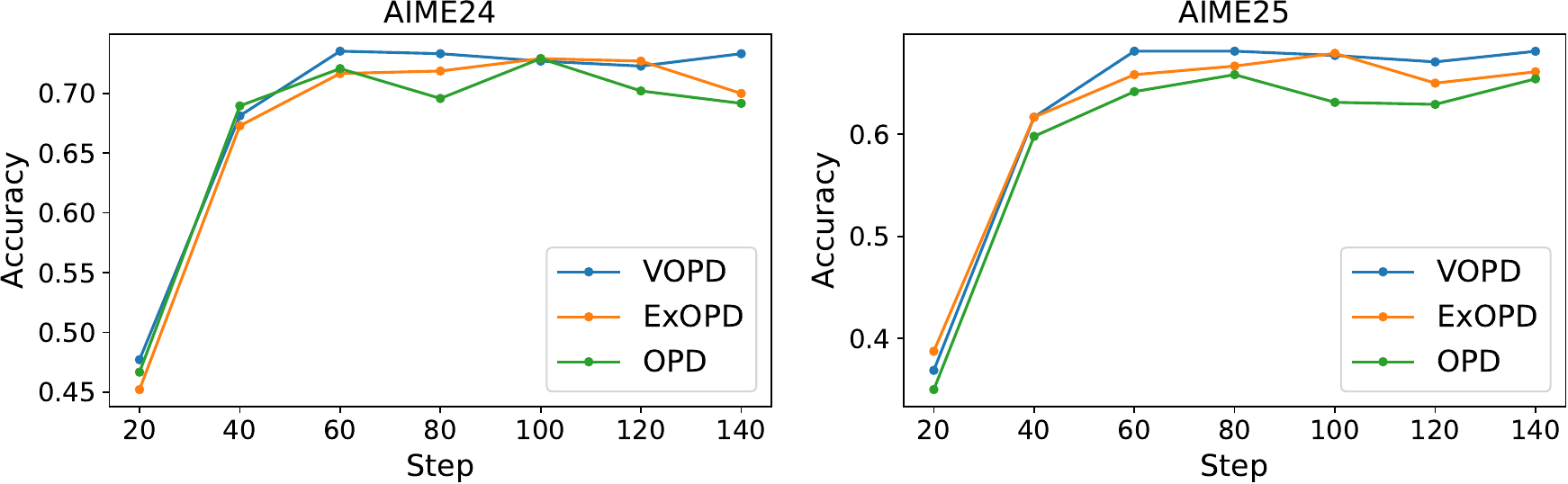}
\captionof{figure}{
Validation accuracy over training in the 4B same-size distillation setting.
Compared with vanilla OPD and fixed-extrapolation ExOPD,
\method improves more rapidly during training and reaches higher converged accuracy.
}
\label{fig:val-curve-4b}
\end{minipage}
\hfill
\begin{minipage}[t]{0.32\linewidth}
\vspace{5pt}
\centering
\resizebox{\linewidth}{!}{
\begin{tabular}{c|cc}
\toprule
$\lambda$ & AIME24 & AIME25 \\
\midrule
-0.25  & 70.02 & 65.42 \\
0.00  & 69.58 & 65.83 \\
0.25 & 73.33 & 68.12 \\
0.50  & 72.91 & 67.91 \\
0.75  & 71.88 & 66.67 \\
\bottomrule
\end{tabular}
}
\vspace{9pt}
\captionof{table}{
Effect of the verifier-alignment strength $\lambda$
in the 4B same-size setting.
}
\label{tab:lambda_ablation}
\end{minipage}

\end{figure*}

\textbf{Effect of Verifier-Grounded Extrapolation Strength $\lambda$.}
We ablate the extrapolation scale \(\lambda\) in the 4B same-size setting on the AIME benchmarks. As shown in~\Cref{tab:lambda_ablation}, \(\lambda=0.25\) performs best on both AIME 2024 and AIME 2025, substantially improving over ordinary OPD (\(\lambda=0\)). Increasing \(\lambda\) further gradually degrades performance. This phenomenon can be explained by the adaptive-extrapolation view from which \(\lambda\) controls how far verifier agreement can move the target beyond the teacher distribution: a moderate value effectively exploits the verifier-grounded teacher signal, whereas an overly large value extrapolates too aggressively and over-amplifies teacher-preferred modes.

\textbf{Analysis of Teacher--Verifier Agreement.} 
A central premise of \method is that the teacher's implicit reward captures part, but not all, of the verifier's preference over candidate responses. If the two signals were unrelated, the projected coefficient $c_x$ would concentrate near zero, and little verifier information could be transferred through the teacher-reward direction. Conversely, if they were perfectly agreed across prompt groups, a fixed extrapolation strength would largely suffice. We therefore examine the empirical agreement coefficient $c_x=\mathrm{Corr}(V,D)$ throughout training.

\begin{figure*}[!t]
\centering
\includegraphics[width=0.99\linewidth]{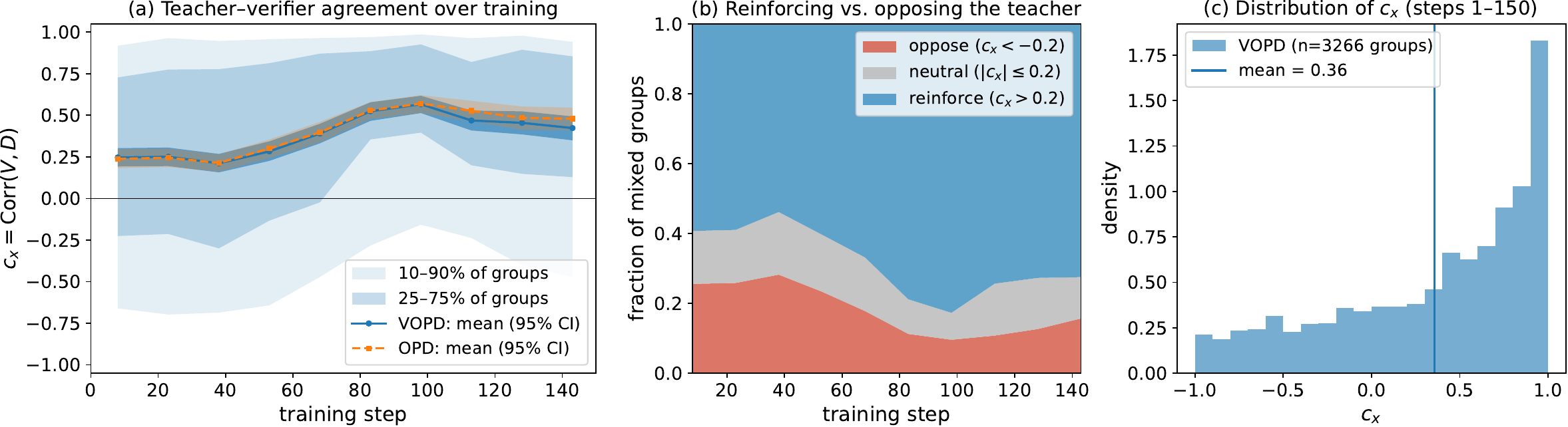}
\caption{\textbf{Analysis of teacher--verifier agreement in practice.}
(a) Mean of agreement coefficient $c_x$ at each training step for \method and vanilla OPD.
(b) Fraction of prompt groups that reinforce, remain neutral to, or oppose the teacher-reward direction over training.
(c) Distribution of $c_x$ across the entire training.}
\label{fig:cx}
\end{figure*}

We study the strong-to-weak distillation setting and find evidence for both parts of the premise. We first investigate whether the teacher reward contains meaningful information about the verifier. The average $c_x$ remains consistently positive throughout training~\Cref{fig:cx}, indicating that the teacher's relative preferences, on average, agree with verifier outcomes. We observe the same pattern under vanilla OPD, showing that this general agreement is already present in ordinary distillation trajectories rather than being created by \method.

At the same time, this agreement is far from uniform across prompt groups. As shown in panels (b)--(c) of~\Cref{fig:cx}, the estimated $c_x$ values are broadly distributed: while most groups reinforce the teacher direction, a nontrivial fraction are approximately neutral or opposing. This means that, in practice, \method does not simply extrapolate positively everywhere: it identifies groups where teacher and verifier preferences disagree and correspondingly weakens or reverses the extrapolation. This adaptive behavior may help explain its performance advantage over fixed-extrapolation methods such as ExOPD.

\begin{figure*}[!t]
  \centering
  \begin{minipage}[t]{0.66\linewidth}
    \vspace{0pt}
    \centering
    \includegraphics[width=\linewidth]{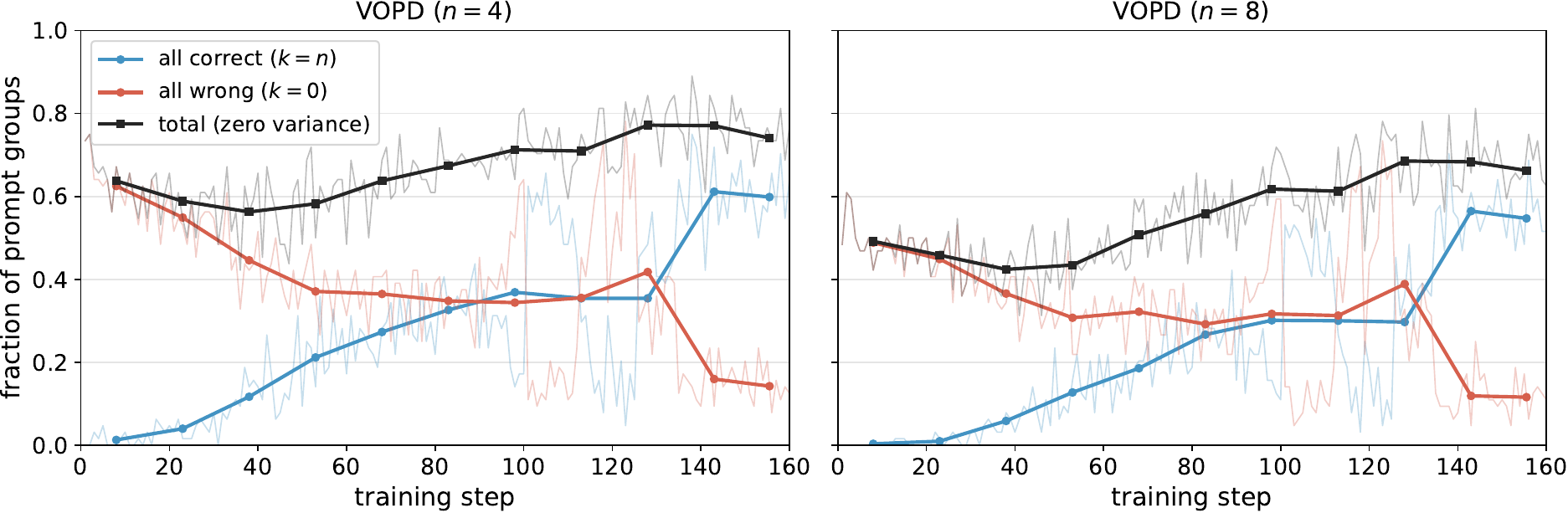}
    \captionof{figure}{\textbf{Zero-variance verifier-reward groups during VOPD training.} Results show fractions of all-correct, all-incorrect, and total zero-variance prompt groups for rollout group sizes $n=4$ (left) and $n=8$ (right).}
    \label{fig:zero_var}
  \end{minipage}
  \hfill
  \begin{minipage}[t]{0.31\linewidth}
    \vspace{20pt}
    \centering
\resizebox{\linewidth}{!}{
  \begin{tabular}{lrr}
    \toprule
    Group size & AIME24 & AIME25 \\
    \midrule
    $n=4$ & \textbf{57.29} & 42.71 \\
    $n=8$ & 57.08 & \textbf{43.54} \\
    \bottomrule
  \end{tabular}
}
\captionof{table}{\textbf{Sensitivity to rollout group size.} Results show VOPD with $n=4$ and $n=8$.}
\label{tab:group_size_sensitivity}

  \end{minipage}
\end{figure*}

\textbf{Zero-variance Groups and Group-size Sensitivity}
To investigate how frequently this fallback occurs and its sensitivity to group size, we additionally evaluate VOPD with $n=8$ under the strong-to-weak setting and compare it with the default $n=4$. \Cref{fig:zero_var} reports the fractions of all-correct, all-incorrect, and total zero-variance groups throughout training. \Cref{tab:group_size_sensitivity} compares downstream performance.

With $n=4$, approximately 65\% of rollout groups have zero verifier-reward variance on average. As training progresses, the fraction of all-incorrect groups decreases while the fraction of all-correct groups increases, leading to an overall increase in zero-variance groups. Increasing the group size to $n=8$ slightly reduces the fraction of zero-variance groups, allowing more groups to provide informative verifier–teacher correlation estimates. Despite this difference, VOPD achieves similar or slightly better performance with $n=8$, suggesting that the method is relatively robust to group size. The modest improvement is consistent with more groups receiving the beneficial verifier-conditioned extrapolation, although other effects of increasing group size cannot be ruled out.

\section{Limitations}
\label{sec:limitations}
Like other forms of on-policy distillation, \method assumes access to a teacher distribution that is meaningfully stronger than the student, which may be unavailable in frontier settings. Recent self-distillation approaches offer a possible alternative through hindsight-informed teacher distributions, and extending \method to such settings is a promising direction. Our method also requires teacher logits to represent verifier feedback through the teacher's dense implicit reward, limiting direct application to closed-source teachers that expose only sampled outputs. Finally, \method requires an additional frozen reference model, increasing memory usage and adding one reference forward pass per rollout batch. The memory cost is inherent to the current formulation, while the additional computation is small relative to rollout generation.

\section{Conclusion}
\label{sec:conclusion}
In this work, we study how to combine the verifier-grounded supervision of RLVR with the dense teacher supervision of on-policy distillation. We introduce \method, which transfers verifier feedback through the dense implicit-reward signal underlying OPD. By placing teacher and verifier feedback in a common reward space, \method projects verifier preference onto the teacher's implicit-reward direction and Rao--Blackwellizes the resulting signal over the full vocabulary, converting sparse verifier feedback into dense, verifier-grounded supervision. Experiments show that \method consistently improves both final performance and learning efficiency over standard OPD and other on-policy distillation baselines, demonstrating its effectiveness in improving language-model reasoning.

\newpage
\section*{AI Assistance Disclosure}
We used generative AI tools to assist with research execution, including method implementation, experimental logging design, and the compilation and visualization of results. We also used generative AI to revise and polish author-written text and mathematical derivations for clarity, grammar, and presentation. Generative AI was not used to retrieve or discover related works, generate synthetic datasets, draft sections of the paper, or prove mathematical claims. All AI-assisted work was reviewed and verified by the authors. The authors take full responsibility for the final content of this work.

\section*{Acknowledgements}
This research was supported by
ARO (W911NF-21-1-0125), ONR (N00014-23-1-2159), the CZ Biohub,
NSF under Nos. CCF-1918940 (Expeditions), DMS-2327709 (IHBEM), IIS-2403318 (III);
NIH under No. 1U24NS146314-01,
Stanford Data Applications Initiative,
Wu Tsai Neurosciences Institute,
Stanford Institute for Human-Centered AI,
Chan Zuckerberg Initiative,
Amazon, Genentech, SAP, and SCBX,

\bibliography{iclr2027_conference}
\bibliographystyle{iclr2027_conference}

\newpage
\appendix
\section{Implementation Details}
\label{app:implementation-details}

\textbf{Training Data.}
All methods are trained on DAPO-Math-17k. Following the prompt alignment used in prior OPD work~\citep{rethinking}, we modify the answer-format instruction to require the final answer in \verb|\boxed{}| form, matching the format used during post-training of the teachers, and adapt the verifier accordingly. The verifier extracts the content of the last \verb|\boxed{}| in the response, normalizes both prediction and ground truth, and checks equivalence by normalized string matching, symbolic equivalence with SymPy, and a \texttt{math\_verify}-based fallback, producing a binary reward in $\{0,1\}$.

\textbf{Additional Training Details.}
We cap prompts at 1,024 tokens and use the response-length limits reported in Table~\ref{tab:hyperparams}; more than 70\% of student rollouts terminate within these limits. Allowing responses to finish is particularly important for \method, since responses truncated before producing a final answer receive a zero verifier signal. After each rollout, the frozen teacher and reference models score the student-generated prefixes by teacher forcing; no teacher-generated trajectories are used during distillation. Following prior OPD implementations, all distillation methods approximate the vocabulary expectation using the student's top-$k$ tokens at each prefix, with $k=16$.

Training uses bfloat16 mixed precision under FSDP. DeepSeek-1.5B weights are loaded in float32, while Qwen3 weights are loaded in bfloat16. Rollout generation and evaluation use vLLM in bfloat16 for all settings.

We train each run until validation accuracy converges. This takes 140 training steps in the 4B same-size setting, and 160 steps in the strong-to-weak setting. These runs all cover less than one pass over DAPO-Math-17k, so no training prompt is encountered twice.

For OPD+GRPO, the sequence-level GRPO advantage is broadcast unchanged to all tokens in the sampled response, with no length normalization or discounting.

\textbf{Hyperparameters.}
Table~\ref{tab:hyperparams} summarizes the training and method-specific hyperparameters for both distillation settings. Unless otherwise specified, we use the same hyperparameters across methods within each setting.

\textbf{Evaluation.} The complete decoding configuration, benchmark sizes, and number of samples are reported in Table~\ref{tab:eval_protocol}. We scale the number of samples per problem $n$ across benchmarks so that each benchmark uses approximately the same amount of evaluation compute.

\begin{table}[h]
\centering
\small
\caption{Evaluation protocol.}
\label{tab:eval_protocol}
\begin{tabular}{lcc}
\toprule
Benchmark & \#Problems & Samples per problem $n$ \\
\midrule
AIME 2024         & 30  & 16 \\
AIME 2025         & 30  & 16 \\
AMC 2023          & 40  & 16 \\
HMMT Feb.\ 2025   & 30  & 16 \\
HMMT Nov.\ 2025   & 30  & 16 \\
MATH-500          & 500 & 4 \\
Minerva Math      & 272 & 4 \\
OlympiadBench     & 674 & 4 \\
\midrule
\multicolumn{3}{l}{Decoding: temperature 0.7, top-$p$ 0.95, max 31{,}744 new tokens, vLLM} \\
\multicolumn{3}{l}{Metric: avg@$n$ (mean accuracy over samples), rule-based \texttt{\textbackslash boxed\{\}} grading} \\
\multicolumn{3}{l}{Checkpoint: final training step} \\
\bottomrule
\end{tabular}
\end{table}

\begin{table}[h]
\centering
\small
\caption{Training hyperparameters for the three distillation settings.
Entries spanning all columns are shared.}
\label{tab:hyperparams}
\begin{tabular}{lcc}
\toprule
Distillation Setting     & 4B Same-size & Strong-to-weak \\
\midrule
Student                  & Qwen3-4B & Qwen3-1.7B \\
Teacher                  & Qwen3-4B-RL-Math & Qwen3-4B-RL-Math \\
Reference $\pi_{\mathrm{ref}}$ & Qwen3-4B & Qwen3-1.7B \\
Training steps           & 140 & 160 \\
\midrule
Thinking mode            & \multicolumn{2}{c}{non-thinking} \\
Group size $G$           & \multicolumn{2}{c}{4} \\
Max response length      & \multicolumn{2}{c}{12{,}288} \\
Weight precision         & \multicolumn{2}{c}{bf16} \\
Training data            & \multicolumn{2}{c}{DAPO-Math-17k (\texttt{\textbackslash boxed\{\}} prompt), 17{,}917 prompts} \\
Batch Size (Prompts per step)         & \multicolumn{2}{c}{64} \\
Minibatch Size           & \multicolumn{2}{c}{64} \\
Max prompt length        & \multicolumn{2}{c}{1{,}024} \\
Rollout temperature / top-$p$ & \multicolumn{2}{c}{1.0 / 1.0} \\
Updates per step / PPO epochs & \multicolumn{2}{c}{1 / 1 (fully on-policy)} \\
Optimizer                & \multicolumn{2}{c}{AdamW, $\beta=(0.9, 0.999)$, wd $=0.01$} \\
Learning rate / schedule & \multicolumn{2}{c}{$2\times10^{-6}$, constant, no warm-up} \\
Gradient clipping        & \multicolumn{2}{c}{1.0} \\
KL penalty / entropy bonus & \multicolumn{2}{c}{none / none} \\
Loss aggregation         & \multicolumn{2}{c}{token-mean} \\
Top-$k$ OPD              & \multicolumn{2}{c}{$k=16$, student top-$k$} \\
Teacher temperature      & \multicolumn{2}{c}{1.0} \\
ExOPD $\lambda$          & \multicolumn{2}{c}{0.25} \\
\method $\lambda$ & \multicolumn{2}{c}{0.25} \\
\bottomrule
\end{tabular}
\end{table}

\end{document}